\documentclass[11pt,a4paper]{article}
\usepackage[hyperref]{acl2017}
\usepackage{times}
\usepackage{latexsym}
\usepackage{url}

\usepackage{amsmath}
\usepackage{amsfonts,amssymb}
\usepackage{graphicx}
\usepackage[T1]{fontenc}
\usepackage{multirow}
\usepackage{wrapfig}

\newcounter{magicrownumbers}
\newcommand\rownumber{\stepcounter{magicrownumbers}\arabic{magicrownumbers}}

\aclfinalcopy 

\title{A Weakly Supervised Approach to Train Temporal Relation Classifiers and Acquire Regular Event Pairs Simultaneously}

\author{Wenlin Yao, Saipravallika Nettyam, Ruihong Huang \\
        Department of Computer Science and Engineering\\
		Texas A\&M University\\
         {\tt \{wenlinyao, n1005120, huangrh\}@tamu.edu}}

\date{}

\begin{document}

\maketitle

\newcommand\SeedAmount{2,000}
\newcommand\BootsAmount{2,000}
\newcommand\TotalAmount{4,401}
\newcommand\BootsThreshold{100}

\begin{abstract}

Capabilities of detecting temporal 
relations between two events can benefit many applications.
Most of existing temporal relation classifiers were trained in a supervised manner.
Instead, we explore the observation that regular event pairs show a consistent temporal relation despite of their various contexts, and these rich contexts can be used to train a contextual temporal relation classifier, which can further recognize new temporal relation contexts and identify new regular event pairs. We focus on detecting \textit{after} and \textit{before} temporal relations and design a weakly supervised learning approach that extracts thousands of regular event pairs and learns a contextual temporal relation classifier simultaneously.
Evaluation shows that the acquired regular event pairs are of high quality and contain rich commonsense knowledge and domain specific knowledge. 
In addition, the weakly supervised trained temporal relation classifier 
achieves comparable performance with the state-of-the-art supervised systems.

\end{abstract}

\section{Introduction}

Capabilities to recognize temporal relations between two events 
can benefit many Natural Language Processing applications, including event timeline generation, script knowledge extraction, text summarization and event prediction. 

This is a challenging task 
because temporal relations can be described in dramatically different contexts depending on domains and pairs of events, signifying different semantic meanings. In order to capture various contexts, large amounts of labeled data are needed to train a high-coverage temporal relation classifier. However, almost all existing datasets that contain event-event temporal relation annotations are limited in size and domains, such as Automatic Context Extraction (ACE) \cite{strassel08} and TimeBank \cite{pustejovsky2003timebank}, which generally contain several hundred documents. 
Most of the existing temporal relation classifiers were trained using these small manually annotated datasets, relying on sophisticated lexical, grammatical, linguistic (e.g., tenses and aspects of events), semantic (e.g., semantic roles and lexicon derived features) and discourse (e.g., temporal discourse connectives \cite{mirza2014classifying}
) features.

We observed that event pairs presenting regularities tend to show the same temporal relation despite of various contexts they may occur in. For instance, {\it arrest} events tend to happen after {\it attack} events, and 
the following sentential contexts all indicate the same temporal relation:

\vspace{.05 in}
\noindent{\it Under pressure following suicide \underline{attacks}, police \underline{arrested} scores of activists on Monday.}

\vspace{.05 in}
\noindent{\it Two men were \underline{arrested} on suspicion of carrying out the Mumbai \underline{attacks}.}

\vspace{.05 in}
\noindent{\it Carlos was \underline{arrested} in Sudan in August in connection with two bomb \underline{attacks} in France in 1982.}

\vspace{.05 in}
\noindent{\it Mamdouh Habib was \underline{arrested} in Pakistan three weeks after the Sept.11 \underline{attacks}.}







Leveraging this key observation, we propose a bootstrapping approach that focuses on recognizing \textit{after} or \textit{before} temporal relations
and substantially reduces the reliance on human annotated data. 
We start by identifying regular event pairs that have occurred enough times with an explicit temporal pattern, i.e., {\small EV\_A} \textit{after (before)} {\small EV\_B}. We then populate these seed event pairs in a large unlabeled corpus
to quickly collect hundreds of thousands of sentences that contain a regular event pair, which are then used as training instances to obtain an initial contextual temporal relation classifier.
Next, the classifier is applied back to the text corpus and label new sentential contexts that indicate a specific \textit{after} or \textit{before} temporal relation between events. 
Then new regular event pairs can be identified, which are event pairs that have a majority of their sentences labeled as describing a particular temporal relation. The newly identified regular event pairs will be used to augment seed event pairs and identify more temporal relation sentential contexts in the unlabeled corpus. The bootstrapping learning process iterates.

In summary, this paper makes the following contributions:
(1) Through this weakly supervised learning method, we obtain both a contextual temporal relation classifier and a list of regular event pairs that usually show a particular "after/before" temporal relation;
(2) Our experiments 
show that the weakly supervised trained contextual temporal relation classifier achieves comparable performance with state-of-the-art supervised models using benchmark evaluation data provided by TempEval-3;
(3) We obtained around 4,400 regular event pairs with the overall accuracy of 69\%. The learned regular event pairs demonstrate rich common sense knowledge, furthermore, our evaluation shows that about 90\% of temporally related regular event pairs are causally related as well.



\section{Related Work}
Most of existing temporal relation classifiers were learned in a supervised manner and depend on human annotated data. In the TempEval campaigns  \cite{Tempeval1,Tempeval2,Tempeval3}, various classification models and linguistic features \cite{Cleartk,chambers2014dense,TIPSem,d2013classifying,mirza2014classifying} have been applied to identify temporal relations between two events.
For example, a recent study by \cite{d2013classifying} applied sophisticated linguistic, semantic and discourse features
to classify temporal relations between events. They also included 437 hand-coded rules in building a hybrid classification model. 
CAEVO, a CAscading EVent Ordering architecture by \citet{chambers2014dense}, applied a sieve-based architecture for event temporal ordering.
CAEVO is essentially a hybrid model as well. While the first few sieves are rule based and  deterministic, the latter ones are machine learned using human annotated data.

In contrast, we present a weakly supervised approach that requires minimal human supervision (i.e., several patterns), 
and simultaneously learns a contextual temporal relation classifier and a collection of regular event pairs. 
In particular, our approach has a co-training \cite{blum1998combining} flavor, 
and the contextual temporal relation classifier learning and the regular event pair acquisition process collaborate and dependent on each other.

Pattern based methods have been applied to acquire event pairs in a specific semantic relation. Specifically, VerbOcean~\cite{VerbOcean} 
extracted fine-gained semantic relations between verbs including the
happens-before relation using lexico-syntactic patterns.
It turns out that the temporal relation patterns used in VerbOcean (e.g., ``to X and then Y'') are too specific and not capable of identifying many event pairs 
that are 
rarely seen in one of the specified patterns.
Evaluation shows that our approach induces very different event pairs from VerbOcean, by 
using the weakly supervised trained temporal relation classifier to recognize diverse contexts that describe a particular temporal relation. 
Our work is also related to previous research on generating narrative event chains \cite{chambers2008unsupervised,chambers2009unsupervised}, however, as indicated by the authors, their focus is not to detect temporal orders between events and the generated event chains are only partially ordered.

Detecting causality between events is challenging and 
has been addressed by several pilot studies
\cite{girju2003automatic,bethard2008learning,riaz2010another,do2011minimally,riaz2013toward}.
Recently, \citet{mirza2014analysis} presented annotation guidelines and annotated explicit causality between events in Timebank. With the resulted corpus, called Causal-TimeBank, they built supervised models to identify causal relations. 
Then \citet{mirza2016catena} proposed a sieve-based method
to perform joint temporal and causal relation extraction, 
exploiting interactions between temporal and causal relations.

\section{Event Representations}
\label{event_representation}



Our bootstrapping approach relies on identifying regular event pairs that tend to unambiguously show a particular temporal relation.
However, an event word can refer to a general type of events or more than one type of events, and therefore has varied meanings depending on contexts. 
To make individual events expressive and self-contained, we 
find and attach arguments to each event word and form event phrases. 
Specifically, we consider both verb event phrases (Section \ref{verb_ep}) and noun event phrases (Section \ref{noun_ep}). We further require that at least one argument is included in an event pair which may be attached to the first or the second event. In other words, we do not consider event pairs in which neither event has an argument.

\subsection{Verb Event Phrases}
\label{verb_ep}
To ensure a good coverage of regular event pairs, we consider all verbs\footnote{We used POS tags to detect verb events.} as event words except reporting verbs\footnote{Reporting verbs,  such as ``said'', ``told'' and ``added'', are commonly seen in news articles. We determined that most of event pairs containing a reporting verb are not very interesting and informative and we therefore discarded these event pairs.}. The thematic patient of a verb refers to the object being acted upon and is essentially part of an event, therefore, we first include the patient of a verb in forming an event phrase. We use Stanford dependency relations \cite{Manning:14} to identify the direct object of an active verb or the subject of a passive verb. 
The agent is also useful to specify a event especially for a intransitive verb event, which does not have a patient. 
Therefore, we include the agent of a verb event in an event phrase if its patient was not found.
Agents are usually the syntactic subject of an active verb or \textit{by} prepositional object of a passive verb. 


For instance, in the sentence \emph{``They win the lottery.''},  
the verb \textit{win} can refer to various \textit{win} events, but with its direct object, \textit{win lottery} refers to a specific type of event.  
For another instance, \emph{``Water evaporates when it's hot.''}, the verb \textit{evaporates} itself is not very meaningful without contexts, but after including its subject, the event \textit{water evaporates} becomes self-contained. If neither a patient nor an agent was found, we include a prepositional direct object of a verb in the event representation to form an event phrase.

\subsection{Noun Event Phrases}
\label{noun_ep}
We include a prepositional object of a noun event in forming an noun event phrase. 
We first consider an object headed by the preposition \textit{of}, then an object headed by the preposition \textit{by}, lastly an object headed by any other preposition.

Note that many noun words do not refer to an event. In order to compile a list of noun event words, we use two intuitive textual patterns \textit{participate in} {\small EVENT} and \textit{involve in} {\small EVENT}.
By the semantics of these two patterns, their prepositional direct objects refer to events. 
However, due to language vagueness and dependency analysis errors, non-event words were seen in the {\small EVENT} position too. 
Therefore, we only consider words that have occurred with one of the two patterns at least 20 times as potential noun event words.
To further remove noise, we quickly went through the list of nouns and manually removed non-event words.
Finally, we obtained 721 noun event words.

\subsection{Generalizing Event Arguments Using Named Entity Types}
\label{generalization}
Including arguments into event representations generates specific event phrases though. 
In order to obtain generalized event phrase forms, we replace specific name arguments with their named entity types~\cite{Manning:14}. 
We also consider replacing pronouns with their types, but concerned with poor quality of full coreference resolution, we only replace personal pronouns with their type {\small PERSON}. We observed that 
this strategy greatly improves generality of event phrases and facilitates the bootstrapping learning process. In section \ref{accuracy}, we compare bootstrapping learning performance using generalized event representations v.s. using non-generalized event representations.

\begin{figure*}[t]
 \centering
 \includegraphics[width = 3.6in]{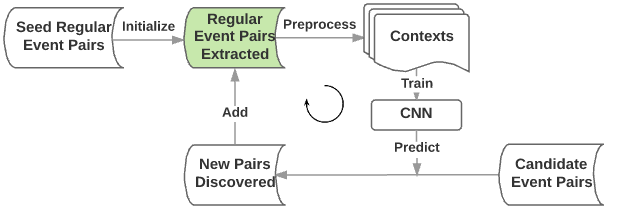}
 \caption{Overview of the Bootstrapping System}
\label{boot_pipeline}
\end{figure*}

\subsection{Regular Event Pair Candidates}
Considering that it is not feasible to test all possible pairs of events in Gigaword and often two events that co-occur in a sentence have no 
temporal relation. In order to narrow down the search space, we identify candidate event pairs which are likely to have temporal relations.

Two strategies are used to identify candidate event pairs. First, by intuition, if two event phrases co-occur (within a sentence) many times, the likelihood of the two events being related and having a temporal relation should be higher compared to event phrases that rarely co-occur.
Therefore, we select event phrase pairs that co-occur within a sentence for more than 100 times as candidate event pairs. 
Second, 
we use two specific temporal relation patterns, {\small EV\_A} \textit{after (before)} {\small EV\_B}, that explicitly indicate two events are in a after (before) relation.
We extract an event pair as a candidate regular pair if
it occurs three or more times with one of the 
patterns in the text corpus. 
The assumption is that if a pair of events shows a particular temporal relation regularly, it is likely to be seen in the above textual patterns as well. 
Specifically, we extract the governor and dependent word of 
the dependency relation {\it prep\_after (prep\_before)} in the annotated English Gigaword \cite{napoles2012annotated} and check whether each word is an event\footnote{Note we consider any verb and a noun that is in our noun event list as an event.}.  
If yes, we form an event phrase for each event 
and obtain an event pair. 
In addition, we expect regular event pairs to occur mostly in a single temporal order, either {\it before} or {\it after}, 
and discard event pairs that have showed mixed temporal orders. 
Specifically, a regular event pair is required to occur in a particular temporal relation  more than 90\% of times. 

Overall by applying the two strategies, we obtained a candidate event pair pool that consists of 40,278 event pairs. 

\section{Bootstrapping both Regular Event Pairs and a Temporal Relation Classifier}

Figure \ref{boot_pipeline} illustrates how the bootstrapping system works.
We first populate seed regular event pairs in the text corpus and identify sentences that contain a regular event pair 
as training instances. We train a contextual temporal relation classifier, using Convolutional Neural Nets (CNNs), to identify specific contexts describing a temporal  \textit{after (before)} relation. We then apply the classifier to the corpus to identify new sentences that describe a particular temporal relation,
from which new regular event pairs can be extracted. Note that the classifier is only applied to sentences that contain a candidate regular event pair.
The bootstrapping process repeats until the number of newly identified regular event pairs is less than 100.



While we used the whole Gigaword \cite{napoles2012annotated} to identify regular event pairs, we only 
use the New York Times section of Gigaword for bootstrapping learning.

\subsection{Regular Event Pair Seeds} \label{Pairs_as_Seeds}

In order to ensure high quality of seed pairs, 
we only consider event pairs that have occurred in explicit temporal relation patterns, {\small EV\_A} \textit{after (before)} {\small EV\_B}, as seed event pairs.
Furthermore, we require each seed regular event pair to have occurred in a temporal relation pattern for at least ten times. 
Specifically, we identified 2,110 seed regular event pairs using the Gigaword corpus\footnote{By populating seed regular event pairs in the New York Times section of the Gigaword corpus, we extracted 7191 sentences and 11339 sentences that contain an event pair in a ``before'' and ``after'' temporal relation respectively.}. 

\subsection{Contextual Temporal Relation Classification}



\begin{figure}[htbp]
 \centering
 \includegraphics[width = 3in]{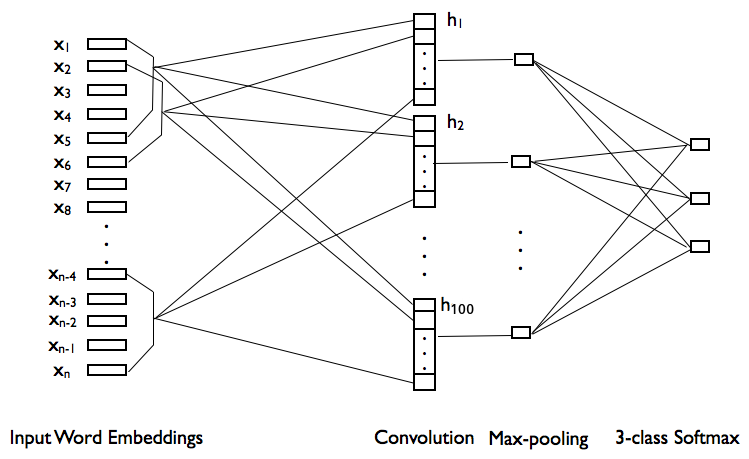}
 \caption{CNN Model Architecture}
\label{CNN}
\end{figure}

We use a neural net classifier to capture compositional meanings of sentential contexts and avoid tedious feature engineering.
Specifically, we used a Convolutional Neural Net (CNN) as our classifier, 
inspired by recent successes of CNN models in various NLP tasks and applications, such as sentiment analysis \cite{Kalch2014,Kim2014}, sequence labeling \cite{Collobert2011} and semantic parsing \cite{Yih2014}.
As shown in figure \ref{CNN}, our CNN architecture is a slight variation of the previous models as described in  \cite{Kim2014,Collobert2011}. 
It has one convolutional layer with 100 hidden nodes, 
one pooling layer and one fully connected softmax layer.

The input are word embeddings of an array of sentential context words.
A convolution filter 
is applied to a sliding window of 
every $h$ words to 
provide input for each hidden node. 
We use Rectified Linear Unit (ReLU) as the non-linear activation function.
%
%
%
We next apply a max-pooling operation to take the maximum value 
over a feature map.
%
The final softmax layer output 
probability distributions over three classes ({\small AFTER}, {\small BEFORE} and {\small OTHER}) indicating the temporal relation between a pair of events in a sentence. Specifically, the temporal relations are defined with respect to the textual order  the two events are presented in a sentence. If the first event is temporally {\small BEFORE} the second event as described in a sentence, this instance will be labeled as {\small BEFORE}. Otherwise if the first event is temporally {\small AFTER} the second event as described in a sentence, the instance will be labeled as {\small AFTER}. The class {\small OTHER} is to capture all the rest contexts that may describe a temporal relation other than \textit{after (before)} or do not describe a temporal relation.

In our experiments, 
we use pre-trained 300-dimention word2vec word embeddings \cite{Mikolov13} that are trained on 100 billion words of Google News and
we use a
filter window size of 5. In training, we used 
stochastic gradient descent
with Adadelta update rule (Zeiler, 2012) and mini-batch size of 100, in addition, we applied dropout \cite{Dropout} with rate $p=0.5$ to avoid overfitting of the CNN model.
We also randomly selected 10\% of the training data as the validation set and chose the classifier with the highest validation performance within the first 10 epochs. 

\subsubsection{Sentential Contexts: Local Windows v.s. Dependency Paths}\label{sentential_contexts}
We explore two types of contexts, local windows v.s. dependency paths, in order to identify contexts that effectively describe temporal relations between two events.

First, the local window based context for an event pair includes five words before the first event, five words after the second event and all the words between the two events. 
Note that two event phrases can be arbitrarily far from each other and long contexts are extremely challenging for a classifier to capture. 
In our experiments, we only consider sentences where two event mentions are at most 10 words away.

Second, we observed that not every word between two events is useful to predict 
their temporal relation. In order to concentrate on relevant context words, 
we further construct dependency path\footnote{Stanford CoreNLP \cite{Manning:14} were used to generate dependency trees.} based context representation.
Specifically, considering a dependency tree as an undirected graph, we use breadth-first-search to extract a sequence of words connecting the first event word to the second event word. In addition, to capture important information in certain syntactic structures such as conjunctions, we extract children nodes for each word in the path. Finally, we sort extracted words according to their textual order in the original sentence and the sorted sequence of words is provided as an input to the CNN classifier.



\subsubsection{Negative Training Instances}\label{negative_instances}
Reasonably, 
most sentences in a corpus do not contain an event pair that is in a temporal ``before/after'' relation.
Therefore, we use negative instances that are 10 times of the total number of positive training instances (i.e., sentences that contain an event pair in a \textit{after (before)} relation). 
Specifically, we require a negative instance to contain an event pair that does not appear in seed pairs nor the candidate event pair set. We randomly sampled negative instances satisfying the condition.
Then these deemed negative instances were labeled as the class {\small OTHER}, a class that compete with the two temporal relation classes, {\small BEFORE} and {\small AFTER}. 




\begin{table*}[h]
\small
\begin{center}
\begin{tabular}{ |l|cccccc|c|}\hline

    Systems & 0 (Seeds) & 1  & 2 & 3 & 4 & 5 &Total\\ \hline
    Basic System &  1057 & 213 & 102 & 48 & -- & -- & 1420  \\ \hline
    + Arg Generalization &  2110 & 638 & 323 & 81 & -- & -- & 3152 \\ \hline
    + Dependency Path Contexts (Full System) &  2110 & 1230 & 555 & 288 & 156 & 62 & 4401  \\ \hline
\end{tabular}
\end{center}
\caption{Number of New Regular Event Pairs Generated after Each Bootstrapping Iteration}
\label{bootstrapping_pairs_num}
\end{table*}

\subsection{New Regular Event Pair Selection Criteria}
Recall that regular event pairs are event pairs that tend 
to show a particular temporal relation despite of their various contexts. 
Therefore, we 
identify a candidate event pair as a new regular event pair if majority of its sentential contexts, specifically 60\% of contexts, were consistently labeled as a particular temporal relation (\textit{after} or \textit{before}) by the CNN classifier.
In addition, we require that at least 15 instances of a regular event pair have been labeled as the majority temporal relation. In order to control semantic drift \cite{mcintosh2009reducing} in bootstrapping learning, we increase the threshold by 5 after each iteration.

Furthermore, in order to filter out ambiguous event pairs that can be in either \textit{before} or \textit{after} temporal order depending on concrete contexts, we require the
absolute difference between number of instances labeled as {\small AFTER} and labeled as {\small BEFORE} to be greater than a ratio of the total number of instances, specifically, we set the ratio to be 40\%.

\section{Evaluation}

\label{sec:evaluation}
Our bootstrapping system learned 
regular event pairs as well as a contextual temporal relation classifier. We evaluate 
each of the two learning outcomes separately.

\subsection{Regular Event Pair Acquisition}

\subsubsection{System Variations}
We compare three variations of our system: 

{\it Basic System}: in the basic system, we did not apply event argument generalization as described in section \ref{generalization}. In addition, we use local window based sentential contexts as input for the classifier.

{\it + Arg Generalization}: on top of the basic system, we apply  event argument generalization.

{\it + Dependency Path Contexts (Full System)}: in the full system, we apply event argument generalization and use dependency path based sentential contexts as input for the classifier.

Table \ref{bootstrapping_pairs_num} shows the number of regular new pairs that were generated after each bootstrapping iteration by each of the three systems. First, we can see that event argument generalization is useful in obtaining roughly two times of seed regular event pairs. Second, event argument generalization is useful in recognizing additional regular event pairs in bootstrapping learning as well. Third, dependency path based sentential contexts are effective in capturing relevant sentential contexts for temporal relation classification, which enables the bootstrapping system to maintain a learning momentum and learn more regular event pairs. 
\subsubsection{Accuracy of Regular Event Pairs}\label{accuracy}

\begin{table}[t]
\small
\centering
\begin{tabular}{|l|l|l|}
\hline
Systems & Seed Pairs & New Pairs \\ \hline
Basic System & 0.73 &  0.55    \\ \hline
+ Arg Generalization  & \multirow{2}{*}{0.71} &  0.63   \\ 
+ Dependency Path Contexts   &       & 0.67  \\ \hline
\end{tabular}
\caption{Accuracy of 100 Randomly Selected Event Pairs}
\label{systems_accuracy}
\end{table}

For each of the three system variations, we randomly selected 50 pairs from seed regular event pairs and 50 from bootstrapped event pairs\footnote{The seed pairs for the second and the third system are the same, so we evaluate the same 50 randomly selected seed pairs for the two systems.} and asked two human annotators to judge the correctness of these acquired regular event pairs. 

Specifically, for each selected event pair, we ask two annotators to label whether a temporal {\small AFTER} or {\small BEFORE} relation exists between the two events.
In addition to the two temporal relation labels, we provide the third category {\small OTHER} as well. We instruct annotators to assign the label {\small OTHER} to an event pair if the two events (i) generally have no temporal relation, (ii) have a temporal relation other than {\small AFTER} or {\small BEFORE}, or (iii) one or both mentions do not refer to an event at all.\footnote{This can happen due to Part-Of-Speech errors or ambiguous event words.} For each event pair, only one label is allowed.
Before the official annotation, we trained the two annotators with system generated event pairs for several iterations. 
The event pairs we used in training annotators are different from the final event pairs we used for evaluation purposes.

Table \ref{systems_accuracy} shows the accuracy of regular event pairs learned by each system variation. 
We determine that an event pair is correctly predicted by a system if the system predicted temporal relation is the same as the label that has been assigned by both of the two annotators. 
The overall kappa inter-agreement between the two annotators is 72\%.
We can see that 
with event argument generalization, the quality of acquired seed regular event pairs is roughly equal to that using specific name arguments. 
Furthermore, because we obtained two times of seed event pairs after using event argument generalization, the second and third bootstrapping systems received more guidance and continued to learn regular event pairs with a high quality. 
In addition, using dependency path based sentential contexts enables the classifier to 
further improve the accuracy of bootstrapped regular event pairs.

\subsubsection{Examples and Constructed Knowledge Graphs}
We have learned around 4,400 regular event pairs that are rich in commonsense knowledge and domain specific knowledge for domains including politics, business, health, sports and crime. 
Table \ref{diff_types} shows several examples in each category. 

\begin{figure}[t]
 \centering
 \includegraphics[width = 3in]{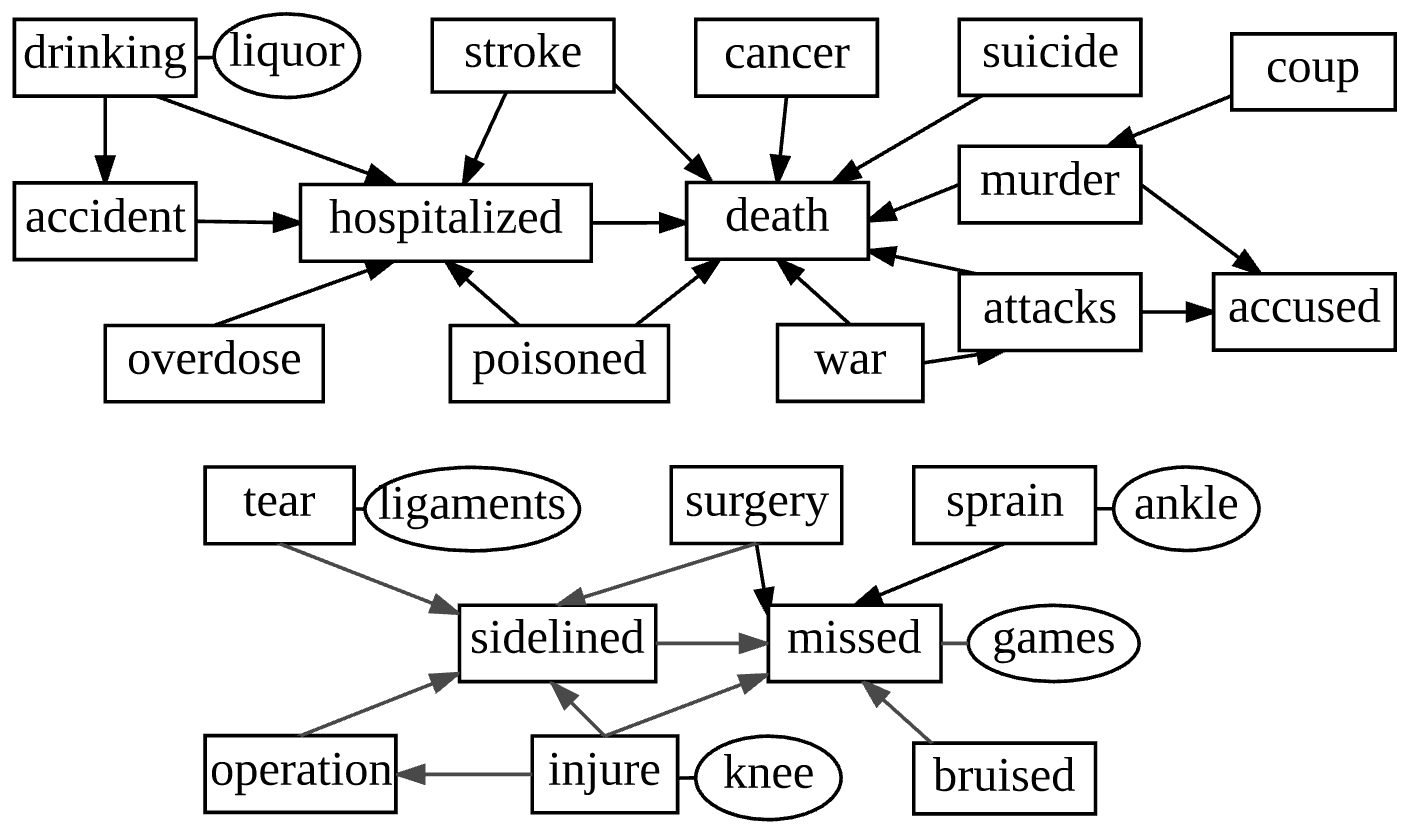}
 \caption{Knowledge Graphs}
\label{knowledge_graph}
\end{figure}

In addition, related event pairs form knowledge graphs, figure~\ref{knowledge_graph} shows two examples. The first one describes various scenarios that cause deaths while the second one describes contingent relations among events specific in sports. 


\begin{table}[t]
\small
\centering
\begin{tabular}[center]{|c|l|} \hline
\multirow{4}{*}{\begin{tabular}[c]{@{}l@{}}Common \\ Sense\end{tabular}}
& PERSON {\bf worked} $\leftarrow$  {\bf graduation} \\
& {\bf career} $\rightarrow$ {\bf announced} retirement \\
& {\bf wash} hands $\rightarrow$ {\bf eating}\\
& PERSON {\bf returned} $\leftarrow$ {\bf visit} \\
\hline
\multirow{4}{*}{\begin{tabular}[c]{@{}l@{}}Politics\end{tabular}} 
& government be {\bf formed} $\leftarrow$ {\bf elections}\\
& {\bf fled} mainland $\leftarrow$ {\bf losing} war \\
& {\bf imposed} sanctions $\leftarrow$ {\bf invasion} of LOCATION \\
& LOCATION {\bf split} $\leftarrow$ {\bf war} \\
\hline
\multirow{3}{*}{\begin{tabular}[c]{@{}l@{}}Business\end{tabular}} 
& {\bf reached} agreement $\leftarrow$ {\bf negotiations}\\
& {\bf hosted} banquet $\leftarrow$ {\bf meeting}\\
& {\bf trading} $\rightarrow$ stock {\bf closed} \\
\hline
\multirow{4}{*}{\begin{tabular}[c]{@{}l@{}}Health\end{tabular}}
& {\bf cause} of death $\leftarrow$ {\bf cancer}\\ 
& PERSON be {\bf hospitalized} $\leftarrow$ {\bf suffering} stroke\\
& PERSON {\bf died} $\leftarrow$ {\bf admitted} to hospital\\
 
\hline
\multirow{4}{*}{\begin{tabular}[c]{@{}l@{}}Sports\end{tabular}}
& {\bf games} $\rightarrow$ {\bf ended} season\\
& PERSON be {\bf sidelined} $\leftarrow$ {\bf undergoing} surgery\\
& PERSON be {\bf suspended} $\leftarrow$ {\bf testing} for cocaine\\
& PERSON {\bf returned} $\leftarrow$ {\bf recovering} from injury \\
\hline
\multirow{4}{*}{\begin{tabular}[c]{@{}l@{}}Crime\end{tabular}}
& {\bf shooting} $\rightarrow$ PERSON be {\bf arrested}\\
& {\bf spending} in jail $\rightarrow$ PERSON be {\bf released}\\
& PERSON be {\bf arrested} $\leftarrow$ {\bf bombings} \\
& driver {\bf fled} $\leftarrow$ {\bf accident} \\
\hline
\end{tabular}
\caption{Examples of Learned Regular Event pairs. 
$\rightarrow$ represents {\it before} relation and $\leftarrow$ represents {\it after} relation.}
\label{diff_types}
\end{table}


\subsubsection{Causally Related Events}
We observed that a large portion of the learned regular event pairs are both temporally and causally related.
We adopt the force dynamics theory and determine that two events are causally related if one event causes, enables or prevents the other event to happen.
Then we asked two annotators \footnote{ We used the same two annotators that have conducted temporal relation annotations. For this task, the annotator inter-agreement is $0.82$ in kappa.} to annotate causal relations for the same set of 100 randomly selected regular event pairs generated by the full bootstrapping system.
Surprisingly, out of 69 event pairs that have been assigned with the same temporal relation by both annotators, 61 event pairs were deemed as causally related.
This shows that most of our temporally related regular event pairs are causally related as well.


\subsubsection{Using VerbOcean Patterns}
\begin{table}[t]
\small
\begin{center}
\begin{tabular}{ |l|cccccc|}\hline

     & 0 (Seeds) & 1  & 2 & 3 & 4 & Total\\ \hline
    Full System &  112 & 179 & 271 & 95 & -- & 657 \\ \hline
\end{tabular}
\end{center}
\caption{Bootstrapping Using VerbOcean Patterns}
\label{bootstrapping_verbocean}
\end{table}

VerbOcean \citet{VerbOcean} created lexico-syntactic patterns in order to extract event pairs with various semantic relations from the Web.  
Specifically, for the temporal relation \textit{happens-before},
VerbOcean used ten patterns such as ``to X and then Y'', ``to X and later Y'' and acquired 4,205 event pairs with a temporal ``before/after'' relation from the Web.

Therefore, we replace our two straightforward temporal relation patterns,  {\small EV\_A} \textit{after (before)} {\small EV\_B}, with the ten patterns proposed by VerbOcean and use these patterns to acquire seed regular event pairs. However, with exactly the same settings and frequency threshold we used in seed identification, we can only identify seven seed regular event pairs using the same complete Gigaword corpus. In order to obtain more seed event pairs, we lowered the frequency threshold of seeing an event pair in patterns from ten to three. In this case as shown in table \ref{bootstrapping_verbocean}, we obtained 112 seed event pairs, which is still much less than 2110 event pairs that we have acquired. 
Then with the initial 112 seed regular event pairs, around 500 new event pairs were later learned using exactly the same bootstrapping learning settings we have used. In total, only 657 event pairs were learned by using VerbOcean patterns.
Note that the Gigaword corpus we used is much smaller in volume than the Web. Therefore, 
we hypothesize that VerbOcean patterns are too specific to be productive in identifying regular event pairs from a limited text corpus.

In addition, we compared our learned 4,401 regular event pairs with the 4,205 verb pairs in the \textit{happens-before} relation acquired by VerbOcean\footnote{Because event pairs in VerbOcean do not contain arguments, we removed event arguments from our event pairs for direct comparisons.}. Interestingly, among these two sets, only eight event pairs are the same. This shows that our bootstrapping learning approach recognizes diverse sentential contexts and learns a dramatically different set of temporally related event pairs, compared with VerbOcean which mainly uses specific lexico-syntactic patterns to query the giant Web.

\subsection{Weakly Supervised Contextual Temporal Relation Classifier}

\subsubsection{Accuracy of the Classifier}
Recall that the contextual temporal relation classifier was trained on the New York Times section of Gigaword. In order to evaluate the accuracy of the classifier, we applied the weakly supervised learned classifier (the full system) to sentential contexts between pairs of events extracted from the Associated Press Worldstream section of Gigaword. 
We randomly sampled 100 instances from the ones that were labeled by the classifier 
as indicating a {\it after} or {\it before} relation and with a confidence score greater than 0.8.
Then for each instance and its pair of events,  
we asked our two annotators to judge whether the sentence 
indeed describes a \textit{after (before)} temporal relation between the two events. 
According to the annotations\footnote{The two annotators achieved a Kappa inter-agreement score of $0.71$.}, the classifier predicted the correct temporal relation 74\% of time.

\subsubsection{Evaluation Using a Benchmark Dataset}
To facilitate direct comparisons, we evaluate both our weakly supervised trained classifier and two supervised trained systems using a benchmark evaluation dataset, the TempEval-3-platinum corpus, which contains 20 news articles annotated with several temporal relations between events. We only evaluate 
system performance on identifying temporal ``before/after'' relations.

We compare with two feature-rich supervised trained systems. 
ClearTK \cite{Cleartk} 
uses event attributes such as tense, aspect and class, dependency paths and words between two events as features in identifying temporal relations between events.
More recently, \cite{mirza2014classifying} proposes even more sophisticated features including various lexical, grammatical and syntactic features, event durations, temporal signals and temporal discourse connectives etc.
In contrast, our neural net based temporal relation classifier is simpler and does not require feature engineering. 

Table \ref{Perf_TempEval} shows the comparison results between these three systems. 
Note that we ran the original ClearTK system and we re-implemented the system described in \cite{mirza2014classifying}.
In addition, both supervised systems were trained using TimeBank v1.2 ~\cite{pustejovsky2006timebank}. 
The performance across the three systems is overall low, one reason is that 
the pairs of events that are in a temporal relation were not provided to the classifiers. Therefore, the classifiers had to identify temporally related event pairs as well as classify their temporal relations.
We can see that the weakly supervised classifier achieved roughly equal performance as ClearTK, while the other supervised system presents a a different precision-recall tradeoff. 
Overall, without using any annotated data 
or sophisticated hand crafted features, our weakly supervised system achieved a F1-score
comparable to both supervised trained systems. 

\begin{table}[t]
\small
\begin{center}
\begin{tabular}{ |c|l|ccc| }
\hline
&Approaches & F1 & P & R \\ \hline
\rownumber & ClearTK~\cite{Cleartk} & 0.27 & 0.36 & 0.22 \\
\rownumber & \citet{mirza2014classifying} & 0.29 & 0.24 & 0.38 \\ \hline
\rownumber & Our classifier & 0.28  & 0.35 & 0.24 \\ \hline

\end{tabular}
\end{center}
\caption{Performance on TempEval-3 Test Data}
\label{Perf_TempEval}
\end{table}

\section{Conclusion}
We presented a weakly supervised bootstrapping approach that learns both regular event pairs and a contextual temporal relation classifier, by exploring the observation that regular event pairs tend to show a consistent temporal relation despite of their diverse contexts.   
Evaluation shows that the learned regular event pairs are of high quality and rich in commonsense knowledge and domain knowledge.
In addition, the weakly supervised trained temporal relation classifier achieves comparable performance with state-of-the-art supervised classifiers. 



\bibliography{acl2017}

\begin{thebibliography}{}
\expandafter\ifx\csname natexlab\endcsname\relax\def\natexlab#1{#1}\fi

\bibitem[{Bethard(2013)}]{Cleartk}
Steven Bethard. 2013.
\newblock Cleartk-timeml: A minimalist approach to tempeval 2013.
\newblock In {\em Second Joint Conference on Lexical and Computational
  Semantics (* SEM)\/}. volume~2, pages 10--14.

\bibitem[{Bethard and Martin(2008)}]{bethard2008learning}
Steven Bethard and James~H Martin. 2008.
\newblock Learning semantic links from a corpus of parallel temporal and causal
  relations.
\newblock In {\em Proceedings of the 46th Annual Meeting of the Association for
  Computational Linguistics on Human Language Technologies: Short Papers\/}.
  Association for Computational Linguistics, pages 177--180.

\bibitem[{Blum and Mitchell(1998)}]{blum1998combining}
Avrim Blum and Tom Mitchell. 1998.
\newblock Combining labeled and unlabeled data with co-training.
\newblock In {\em Proceedings of the eleventh annual conference on
  Computational learning theory\/}. ACM, pages 92--100.

\bibitem[{Chambers et~al.(2014)Chambers, Cassidy, McDowell, and
  Bethard}]{chambers2014dense}
Nathanael Chambers, Taylor Cassidy, Bill McDowell, and Steven Bethard. 2014.
\newblock Dense event ordering with a multi-pass architecture.
\newblock {\em Transactions of the Association for Computational Linguistics\/}
  2:273--284.

\bibitem[{Chambers and Jurafsky(2009)}]{chambers2009unsupervised}
Nathanael Chambers and Dan Jurafsky. 2009.
\newblock Unsupervised learning of narrative schemas and their participants.
\newblock In {\em Proceedings of the Joint Conference of the 47th Annual
  Meeting of the ACL and the 4th International Joint Conference on Natural
  Language Processing of the AFNLP: Volume 2-Volume 2\/}. Association for
  Computational Linguistics, pages 602--610.

\bibitem[{Chambers and Jurafsky(2008)}]{chambers2008unsupervised}
Nathanael Chambers and Daniel Jurafsky. 2008.
\newblock Unsupervised learning of narrative event chains.
\newblock In {\em ACL\/}. Citeseer, volume 94305, pages 789--797.

\bibitem[{Chklovski and Pantel(2004)}]{VerbOcean}
Timothy Chklovski and Patrick Pantel. 2004.
\newblock Verbocean: Mining the web for fine-grained semantic verb relations.
\newblock In {\em Proceedings of Conference on Empirical Methods in Natural
  Language Processing (EMNLP-04)\/}.

\bibitem[{Collobert et~al.(2011)Collobert, Weston, Bottou, Karlen, Kavukcuoglu,
  and Kuksa}]{Collobert2011}
Ronan Collobert, Jason Weston, L{\'e}on Bottou, Michael Karlen, Koray
  Kavukcuoglu, and Pavel Kuksa. 2011.
\newblock Natural language processing (almost) from scratch.
\newblock {\em Journal of Machine Learning Research\/} 12(Aug):2493--2537.

\bibitem[{Do et~al.(2011)Do, Chan, and Roth}]{do2011minimally}
Quang~Xuan Do, Yee~Seng Chan, and Dan Roth. 2011.
\newblock Minimally supervised event causality identification.
\newblock In {\em Proceedings of the Conference on Empirical Methods in Natural
  Language Processing\/}. Association for Computational Linguistics, pages
  294--303.

\bibitem[{D'Souza and Ng(2013)}]{d2013classifying}
Jennifer D'Souza and Vincent Ng. 2013.
\newblock Classifying temporal relations with rich linguistic knowledge.
\newblock In {\em HLT-NAACL\/}. pages 918--927.

\bibitem[{Girju(2003)}]{girju2003automatic}
Roxana Girju. 2003.
\newblock Automatic detection of causal relations for question answering.
\newblock In {\em Proceedings of the ACL 2003 workshop on Multilingual
  summarization and question answering-Volume 12\/}. Association for
  Computational Linguistics, pages 76--83.

\bibitem[{Hinton et~al.(2012)Hinton, Srivastava, Krizhevsky, Sutskever, and
  Salakhutdinov}]{Dropout}
Geoffrey~E Hinton, Nitish Srivastava, Alex Krizhevsky, Ilya Sutskever, and
  Ruslan~R Salakhutdinov. 2012.
\newblock {Improving Neural Networks by Preventing Co-adaptation of Feature
  Detectors}.
\newblock In {\em arXiv preprint arXiv:1207.0580\/}.

\bibitem[{Kalchbrenner et~al.(2014)Kalchbrenner, Grefenstette, and
  Blunsom}]{Kalch2014}
Nal Kalchbrenner, Edward Grefenstette, and Phil Blunsom. 2014.
\newblock A convolutional neural network for modelling sentences.
\newblock In {\em Proceedings of the 52nd Annual Meeting of the Association for
  Computational Linguistics\/}.

\bibitem[{Kim(2014)}]{Kim2014}
Yoon Kim. 2014.
\newblock {Convolutional neural networks for sentence classification}.
\newblock In {\em Proceedings of 2014 the Conference on Empirical Methods in
  Natural Language Processing (EMNLP-2014)\/}.

\bibitem[{Llorens et~al.(2010)Llorens, Saquete, and Navarro}]{TIPSem}
Hector Llorens, Estela Saquete, and Borja Navarro. 2010.
\newblock Tipsem (english and spanish): Evaluating crfs and semantic roles in
  tempeval-2.
\newblock In {\em Proceedings of the 5th International Workshop on Semantic
  Evaluation\/}. Association for Computational Linguistics, pages 284--291.

\bibitem[{Manning et~al.(2014)Manning, Surdeanu, Bauer, Finkel, Bethard, and
  McClosky}]{Manning:14}
Christopher~D. Manning, Mihai Surdeanu, John Bauer, Jenny Finkel, Steven~J.
  Bethard, and David McClosky. 2014.
\newblock The stanford corenlp natural language processing toolkit.
\newblock In {\em Proceedings of the 52nd Annual Meeting of the Association for
  Computational Linguistics (ACL)\/}. pages 55--60.

\bibitem[{McIntosh and Curran(2009)}]{mcintosh2009reducing}
Tara McIntosh and James~R Curran. 2009.
\newblock Reducing semantic drift with bagging and distributional similarity.
\newblock In {\em Proceedings of the Joint Conference of the 47th Annual
  Meeting of the ACL and the 4th International Joint Conference on Natural
  Language Processing of the AFNLP: Volume 1-Volume 1\/}. Association for
  Computational Linguistics, pages 396--404.

\bibitem[{Mikolov et~al.(2013)Mikolov, Sutskever, Chen, Corrado, and
  Dean}]{Mikolov13}
Tomas Mikolov, Ilya Sutskever, Kai Chen, Greg~S Corrado, and Jeff Dean. 2013.
\newblock Distributed representations of words and phrases and their
  compositionality.
\newblock In {\em Advances in neural information processing systems\/}. pages
  3111--3119.

\bibitem[{Mirza and Tonelli(2014{\natexlab{a}})}]{mirza2014analysis}
Paramita Mirza and Sara Tonelli. 2014{\natexlab{a}}.
\newblock An analysis of causality between events and its relation to temporal
  information.
\newblock In {\em COLING\/}. pages 2097--2106.

\bibitem[{Mirza and Tonelli(2014{\natexlab{b}})}]{mirza2014classifying}
Paramita Mirza and Sara Tonelli. 2014{\natexlab{b}}.
\newblock Classifying temporal relations with simple features.
\newblock In {\em EACL\/}. volume~14, pages 308--317.

\bibitem[{Mirza and Tonelli(2016)}]{mirza2016catena}
Paramita Mirza and Sara Tonelli. 2016.
\newblock Catena: Causal and temporal relation extraction from natural language
  texts.
\newblock In {\em The 26th International Conference on Computational
  Linguistics\/}. pages 64--75.

\bibitem[{Napoles et~al.(2012)Napoles, Gormley, and
  Van~Durme}]{napoles2012annotated}
Courtney Napoles, Matthew Gormley, and Benjamin Van~Durme. 2012.
\newblock Annotated gigaword.
\newblock In {\em Proceedings of the Joint Workshop on Automatic Knowledge Base
  Construction and Web-scale Knowledge Extraction\/}. Association for
  Computational Linguistics, pages 95--100.

\bibitem[{Pustejovsky et~al.(2003)Pustejovsky, Hanks, Sauri, See, Gaizauskas,
  Setzer, Radev, Sundheim, Day, Ferro et~al.}]{pustejovsky2003timebank}
James Pustejovsky, Patrick Hanks, Roser Sauri, Andrew See, Robert Gaizauskas,
  Andrea Setzer, Dragomir Radev, Beth Sundheim, David Day, Lisa Ferro, et~al.
  2003.
\newblock The timebank corpus.
\newblock In {\em Corpus linguistics\/}. volume 2003, page~40.

\bibitem[{Pustejovsky et~al.(2006)Pustejovsky, Verhagen, Saur{\'\i}, Littman,
  Gaizauskas, Katz, Mani, Knippen, and Setzer}]{pustejovsky2006timebank}
James Pustejovsky, Marc Verhagen, Roser Saur{\'\i}, Jessica Littman, Robert
  Gaizauskas, Graham Katz, Inderjeet Mani, Robert Knippen, and Andrea Setzer.
  2006.
\newblock Timebank 1.2.
\newblock {\em Linguistic Data Consortium\/} 40.

\bibitem[{Riaz and Girju(2010)}]{riaz2010another}
Mehwish Riaz and Roxana Girju. 2010.
\newblock Another look at causality: Discovering scenario-specific contingency
  relationships with no supervision.
\newblock In {\em Semantic Computing (ICSC), 2010 IEEE Fourth International
  Conference on\/}. IEEE, pages 361--368.

\bibitem[{Riaz and Girju(2013)}]{riaz2013toward}
Mehwish Riaz and Roxana Girju. 2013.
\newblock Toward a better understanding of causality between verbal events:
  Extraction and analysis of the causal power of verb-verb associations.
\newblock In {\em Proceedings of the annual SIGdial Meeting on Discourse and
  Dialogue (SIGDIAL)\/}. Citeseer.

\bibitem[{Strassel et~al.(2008)Strassel, Przybocki, Peterson, Song, and
  Maeda}]{strassel08}
Stephanie Strassel, Mark~A Przybocki, Kay Peterson, Zhiyi Song, and Kazuaki
  Maeda. 2008.
\newblock {Linguistic Resources and Evaluation Techniques for Evaluation of
  Cross-Document Automatic Content Extraction}.
\newblock In {\em Proceedings of the Sixth International Language Resources and
  Evaluation Conference (LREC-08)\/}.

\bibitem[{UzZaman et~al.(2013)UzZaman, Llorens, Allen, Derczynski, Verhagen,
  and Pustejovsky}]{Tempeval3}
Naushad UzZaman, Hector Llorens, James Allen, Leon Derczynski, Marc Verhagen,
  and James Pustejovsky. 2013.
\newblock {SemEval-2013 task 1: TempEval-3 evaluating time expressions, events,
  and temporal relations}.
\newblock In {\em Proceedings of the 7th International Workshop on Semantic
  Evaluation (SemEval 2013)\/}.

\bibitem[{Verhagen et~al.(2007)Verhagen, Gaizauskas, Schilder, Hepple, Katz,
  and Pustejovsky}]{Tempeval1}
Marc Verhagen, Robert Gaizauskas, Frank Schilder, Mark Hepple, Graham Katz, and
  James Pustejovsky. 2007.
\newblock Semeval-2007 task 15: Tempeval temporal relation identification.
\newblock In {\em Proceedings of the 4th International Workshop on Semantic
  Evaluations\/}. Association for Computational Linguistics, pages 75--80.

\bibitem[{Verhagen et~al.(2010)Verhagen, Sauri, Caselli, and
  Pustejovsky}]{Tempeval2}
Marc Verhagen, Roser Sauri, Tommaso Caselli, and James Pustejovsky. 2010.
\newblock Semeval-2010 task 13: Tempeval-2.
\newblock In {\em Proceedings of the 5th international workshop on semantic
  evaluation\/}. Association for Computational Linguistics, pages 57--62.

\bibitem[{Yih et~al.(2014)Yih, He, and Meek}]{Yih2014}
Wen{-}tau Yih, Xiaodong He, and Christopher Meek. 2014.
\newblock Semantic parsing for single-relation question answering.
\newblock In {\em Proceedings of the 52nd Annual Meeting of the Association for
  Computational Linguistics\/}.

\end{thebibliography}
\bibliographystyle{acl_natbib}

\end{document}